\documentclass[]{fairmeta}

\usepackage{amsmath}
\usepackage{amssymb}
\usepackage{amsfonts}
\usepackage{mathtools}
\usepackage{amsthm}
\usepackage{comment}
\usepackage{siunitx}
\usepackage{pifont}
\usepackage{adjustbox}
\usepackage{makecell}
\usepackage{wrapfig}
\usepackage{algorithm}
\usepackage{algorithmicx}
\usepackage{algpseudocode}
\usepackage{float}
\usepackage{listings}
\usepackage{manyfoot}
\usepackage[title]{appendix}
\usepackage{mathrsfs}
\usepackage{colortbl}
\usepackage{tabularx}
\usepackage{enumitem}
\usepackage{dblfloatfix}
\usepackage{capt-of}
\usepackage{nicefrac}
\usepackage{multicol}

\newcommand{\openvla}{\textsc{OpenVLA}}
\newcommand{\basevla}{\textsc{BaseVLA}}
\newcommand{\fullsoft}{\textsc{FullSoft}}
\newcommand{\dynamicsoft}{\textsc{VisualThink-VLA}}
\newcommand{\method}{\dynamicsoft}
\newcommand{\thinkimg}{\emph{think with images}}

\newcommand{\routeyes}{\textcolor{green!50!black}{\scriptsize\ding{51}}}
\newcommand{\routeno}{\textcolor{red!75!black}{\scriptsize\ding{55}}}

\definecolor{tblheader}{RGB}{233,238,245}
\definecolor{tblsubheader}{RGB}{245,247,250}
\definecolor{tblours}{RGB}{238,247,238}
\definecolor{tblreported}{RGB}{244,246,250}
\definecolor{tblna}{gray}{0.55}
\definecolor{tblgrouptext}{RGB}{248,242,232}
\definecolor{tblgrouprl}{RGB}{235,242,249}
\definecolor{tblgroupvisual}{RGB}{243,239,248}
\definecolor{tblgroupmatched}{RGB}{242,244,248}
\newcommand{\hdr}[1]{\colorbox{tblheader}{\strut #1}}

\theoremstyle{plain}

\theoremstyle{definition}

\theoremstyle{remark}

\title{%
\raisebox{-0.3ex}{\includegraphics[height=0.34in]{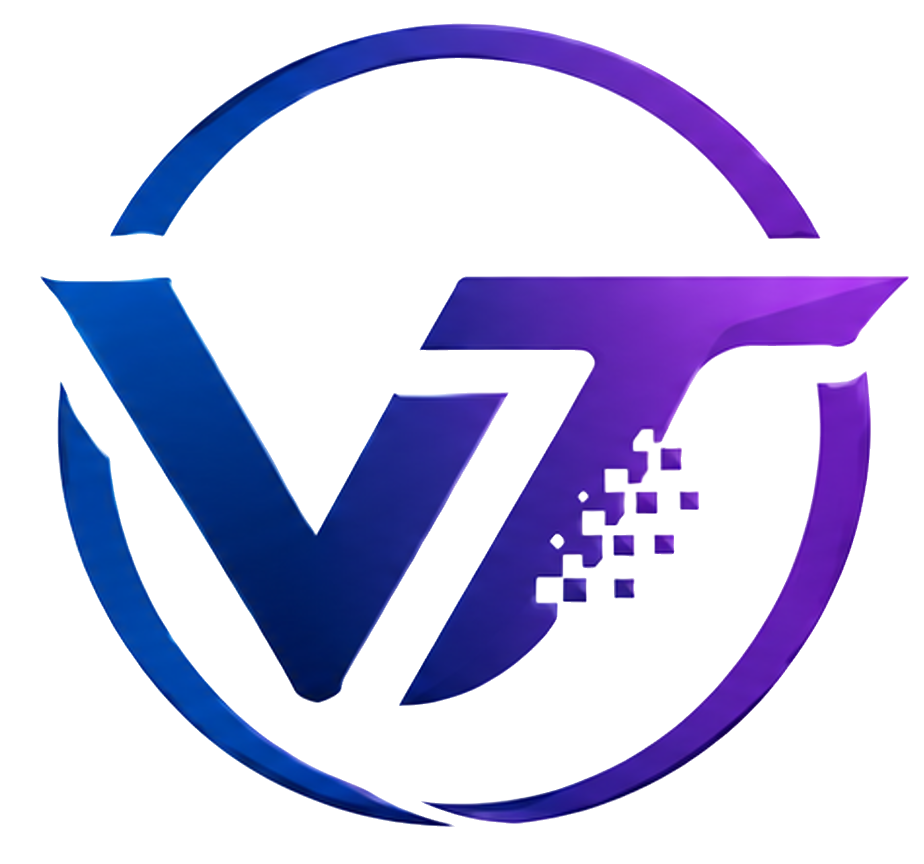}}%
\hspace{0.2em}%
\raisebox{-0.3ex}{\includegraphics[height=0.28in,width=2.0in]{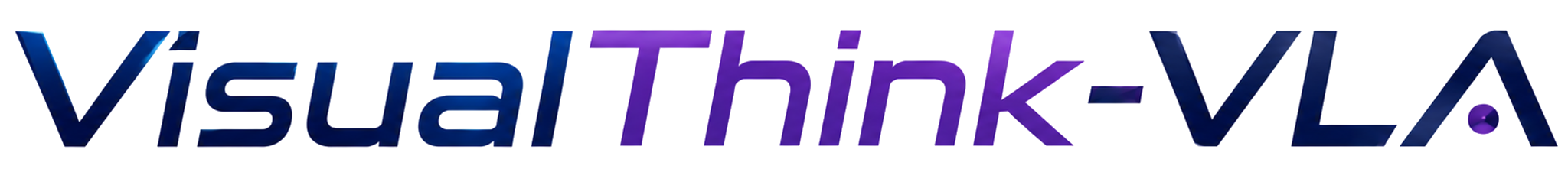}}:
Visual Intermediate Reasoning for Effective and Low-Latency Vision-Language-Action Policies}

\author[1]{Mingjian Gao}
\author[1,\ddagger]{Wenqiao Zhang}
\author[1]{Yuqian Yuan}
\author[1]{Yang Dai}
\author[1]{Binhe Yu}
\author[2]{Zheqi Lv}
\author[1]{Haoyu Zheng}
\author[3]{Jiaqi Zhu}
\author[1]{Zhiqi Ge}
\author[4]{Zixuan Wan}
\author[1]{Siliang Tang}
\author[1]{Yueting Zhuang}

\affiliation[1]{Zhejiang University}
\affiliation[2]{Cornell University}
\affiliation[3]{National University of Singapore}
\affiliation[4]{Xi'an University of Electronic Science and Technology}

\contribution[\ddagger]{Corresponding author}

\abstract{Recent work has begun to equip vision-language-action (VLA) policies with explicit intermediate reasoning. In embodied control, however, textual chain-of-thought is a poor fit: irrelevant or weakly textual information can interfere with action prediction, while autoregressive text decoding adds too much latency for real-time closed-loop execution.
We present \textbf{\method{}}, a visual intermediate-reasoning framework for accurate, low-latency VLA policies.
Our bootstrapping philosophy is to \emph{guide action with effective visual thinking}: \method{} bootstraps action prediction through a compact visual-evidence interface that preserves spatial precision while avoiding decoding overhead.
Besides, to further improve performance and efficiency, \method{} adopts a tailored selective routing mechanism to learn the visual evidence tokens, enabling low-latency inference while preserving high-capacity specialization. We also introduce \textbf{\texttt{VisualEvidence-Kit}}, a supervision-and-audit resource centered on a \textbf{\texttt{VisualEvidence-Agent}} that constructs a 754.7k VLA instructions \textbf{\texttt{VisualEvidence-Set}} for route supervision and counterfactual faithfulness tests.
Across multiple benchmarks and real-robot evaluation, \method{} achieves the highest success rate on most benchmarks while reducing the \textbf{multi-second} latency of reasoning-augmented baselines to the \textbf{sub-second} regime. For example, on BridgeData V2, it reduces step latency from 8.377\,s with ECoT to 0.367\,s, achieving a \textbf{22.8$\times$ speedup}.

\par\smallskip\noindent
\renewcommand{\arraystretch}{1.1}
\begin{tabular}{rl}
    \raisebox{-0.2em}{\includegraphics[height=1.1em]{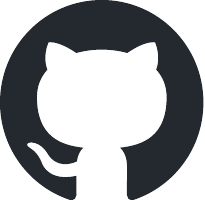}} &
    \textbf{GitHub:} \url{https://github.com/DCDmllm/VisualThink-VLA}
\end{tabular}
}
\begin{document}

\maketitle

\section{Introduction}

Vision-language-action (VLA) policies map visual observations and language instructions directly to robot actions, enabling general-purpose language-conditioned control across diverse embodiments and tasks \citep{brohan2023rt2,kim24openvla,octo2024,black2024pi0}. Yet direct action prediction can be brittle in manipulation scenarios that require resolving distractors, grounding spatial relations, tracking motion, or maintaining progress over long horizons. To address these challenges, recent work augments VLA policies with intermediate thinking before action prediction, improving robustness under spatial ambiguity and task-level constraints \citep{zawalski2024ecot,duan2025fastecot,chen2025cotvla,zheng2024tracevla,qu2025spatialvla,chen2025internvlam1,yin2025deepthinkvla}.

Despite their promise, existing reasoning-augmented VLA policies expose an \textbf{accuracy-efficiency trade-off}. Textual chain-of-thought provides an explicit reasoning trace, but its weak visual grounding can distract action prediction and autoregressive decoding incurs substantial latency for closed-loop control \citep{wei2022chain,kojima2022large,jacovi2020towards}. Dense visual or spatial side information stays closer to the scene, but may also introduce interference: redundant channels can overwhelm the action decoder, irrelevant cues can compete with task-critical evidence, and noisy auxiliary perception can propagate conflicting signals. Thus, more reasoning context is not always better for real-time control.

This motivates our central goal: identifying a \emph{minimal yet effective visual reasoning interface} for VLA policies. Here, ``Minimal'' means avoiding long text traces and dense always-on side information, while ``effective'' means preserving the visual structure required for action. Our premise is that embodied reasoning should remain grounded in visual space while exposing only decision-relevant evidence. This is consistent with recent visual-planning results showing that image-based intermediate reasoning can be better suited than text-space planning for spatial decisions \citep{xu2025visual}.

\begin{figure}[H]
\centering
\includegraphics[width=\textwidth]{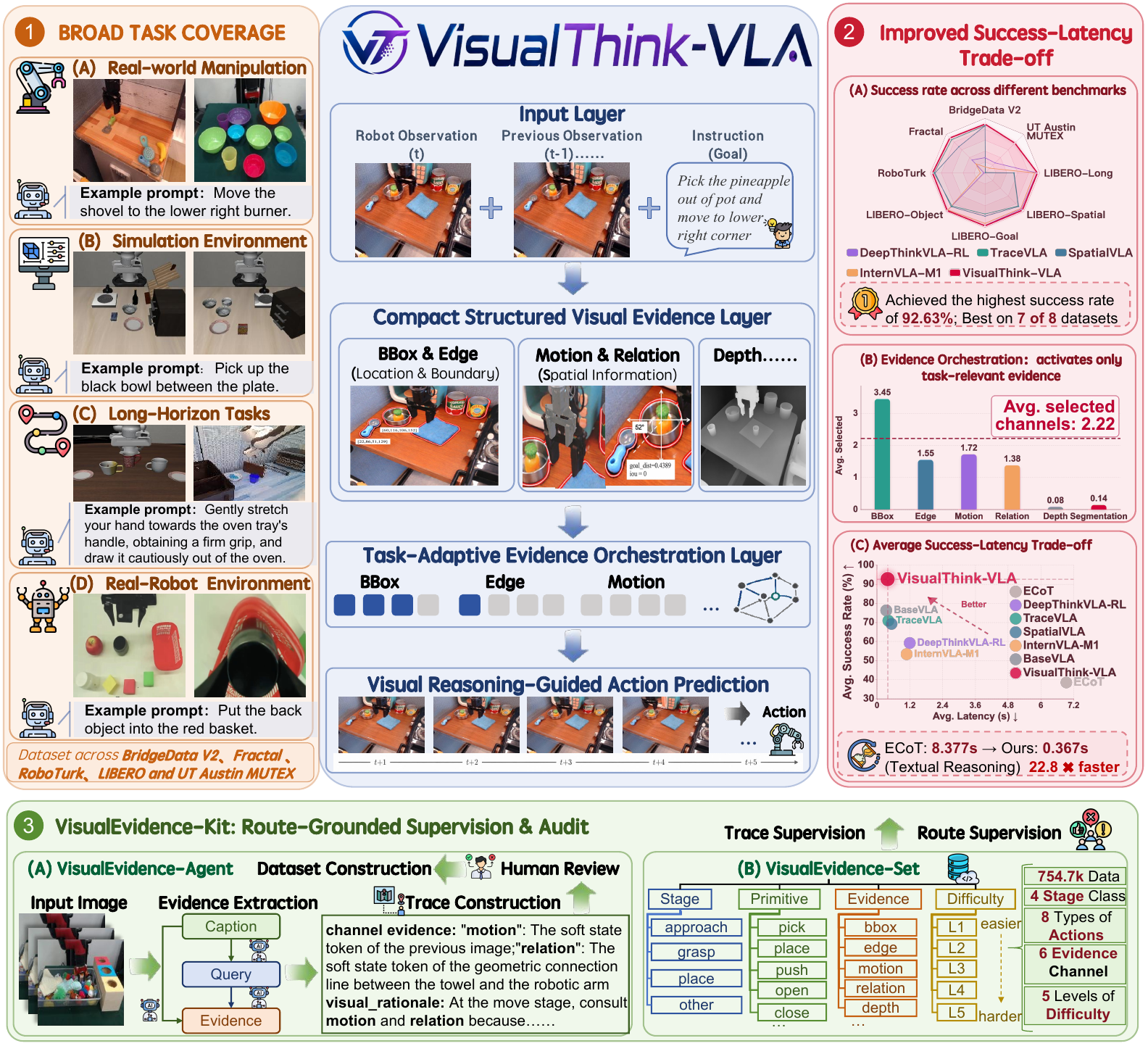}
\caption{Overview of \method{}, covering its task coverage, the success-latency trade-off against prior VLA reasoning baselines, and the route-grounded supervision asset built with VisualEvidence-Kit.}
\label{fig:overview}
\end{figure}

To turn this motivation into a control interface, we propose \textbf{\method{}} (see \cref{fig:overview}), a sparse visual intermediate-reasoning framework that enables a frozen VLA backbone to \thinkimg{} through routed visual evidence. \method{} serves as a general plug-and-play visual reasoning module for VLA policies, conditioning action decoding with lightweight routed evidence states while leaving the underlying backbone unchanged.
\method{} first constructs a compact six-channel evidence layer, then uses a task-adaptive orchestration layer to select useful channels at each decision step.  The selected evidence is injected as learned soft states before action decoding, allowing the policy to remain visually grounded while filtering unnecessary or distracting cues and avoiding the latency of autoregressive textual reasoning.

To preserve the benefit of richer evidence under sparse inference, \method{} couples task-adaptive routing with dense-to-sparse optimization.
A task-adaptive router selects which channels to consult, and a routed adapter conditions the frozen backbone on the selected cues, following the broader principle of conditional computation \citep{shazeer2017moe,fedus2022switch}. To stabilize sparse routing, we train the adapter with soft-hard collaborative masks and distillation from \textbf{\fullsoft{}}, a dense teacher that consults all four screened effective evidence channels \citep{hinton2015distilling}. The four-channel teacher is obtained after channel screening removes two low-utility channels, which were found to contribute little utility while increasing latency and interference risk. The sparse policy therefore inherits much of the dense teacher's benefit while reducing active evidence and exposure to irrelevant cues.

Because routed evidence should be inspectable rather than merely latent, we further introduce \textbf{VisualEvidence-Kit} for supervision and audit. At its core, a \textbf{VisualEvidence-Agent} extracts channel-level evidence, constructs channel-grounded traces, and incorporates human review to build \textbf{VisualEvidence-Set}. 
The resulting VisualEvidence-Set contains 754.7k visual-thinking VLA instructions, forming a route-grounded collection that covers task stages, action primitives, evidence dependence, difficulty, and channel-level targets.
This resource supports route supervision, trace supervision, and counterfactual faithfulness checks without relying on free-form rationales \citep{deyoung2020eraser,jacovi2020towards}.

Finally, we evaluate whether this interface improves control while keeping reasoning lightweight. Experiments across real-world manipulation, simulation, long-horizon tasks, and closed-loop robot execution demonstrate that \method{} improves the success-latency trade-off over reasoning-augmented VLA baselines. It brings textual-reasoning latency back to the sub-second control regime, preserves nearly the performance of dense visual evidence while selecting only a small number of channels, and yields routed evidence that is stage-sensitive and behaviorally aligned with policy decisions.

The main contributions are threefold:
\begin{itemize}[itemsep=0mm,leftmargin=4mm]
    \item \textbf{A visual intermediate-reasoning interface for frozen VLAs.} We formulate \thinkimg{} as action decoding conditioned on compact routed visual cues rather than long text traces or dense auxiliary perception.
    \item \textbf{A task-adaptive evidence orchestration mechanism.} \method{} combines channel screening, sparse evidence routing, soft-hard collaborative optimization, and teacher-student distillation to improve the success-latency trade-off while filtering irrelevant channels.
    \item \textbf{A route-grounded supervision and audit framework.} \textbf{VisualEvidence-Kit} centers on VisualEvidence-Agent, which performs evidence extraction, trace construction, and human review to build VisualEvidence-Set for faithfulness diagnostics.
\end{itemize}


\section{Related Work}

\noindent\textbf{VLA Reasoning Policies.}
Large-scale robot policies increasingly use language-conditioned sequence models trained on heterogeneous demonstration corpora. Representative VLA systems include \openvla{}, Octo, GR-2, and $\pi_0$ \citep{kim24openvla,octo2024,cheang2024gr2,black2024pi0}, building on broader language-conditioned robot learning over corpora such as Open X-Embodiment and RT-X \citep{openx2023,ahn2022saycan,driess2023palm,brohan2023rt2}. Recent reasoning-augmented VLA methods add explicit intermediate reasoning or stronger grounding, including ECoT for textual reasoning traces and TraceVLA or SpatialVLA for visual or spatial grounding \citep{zawalski2024ecot,zheng2024tracevla,qu2025spatialvla}. These methods show that reasoning can help control, but they also motivate our focus on a minimal visual interface: textual CoT introduces closed-loop latency, while dense visual or spatial side information can introduce irrelevant or redundant evidence.

\begin{table}[H]
\centering
\small
\setlength{\tabcolsep}{7pt}
\renewcommand{\arraystretch}{1.08}
\resizebox{\textwidth}{!}{%
\begin{tabular}{@{}lcccccc@{}}
\toprule
\textbf{Method} & \textbf{Location} & \textbf{Boundary / Geometry} & \textbf{Motion History} & \textbf{Relation} & \textbf{Depth} & \textbf{Segmentation} \\
\midrule
\openvla{} \citep{kim24openvla} & \routeno{} & \routeno{} & \routeno{} & \routeno{} & \routeno{} & \routeno{} \\
ECoT \citep{zawalski2024ecot} & \routeyes{} & \routeno{} & \routeno{} & \routeno{} & \routeno{} & \routeno{} \\
TraceVLA \citep{zheng2024tracevla} & \routeno{} & \routeno{} & \routeyes{} & \routeno{} & \routeno{} & \routeno{} \\
SpatialVLA \citep{qu2025spatialvla} & \routeno{} & \routeyes{} & \routeno{} & \routeno{} & \routeyes{} & \routeno{} \\
\rowcolor{tblours}
\method{} & \routeyes{} & \routeyes{} & \routeyes{} & \routeyes{} & \routeyes{} & \routeyes{} \\
\bottomrule
\end{tabular}
}
\caption{Explicit visual cues used by representative VLA methods.Depth and segmentation are included in the initial candidate bank but screened out from the default operational setting.}
\label{tab:visual_cue_compare}
\end{table}

\noindent\textbf{Visual Evidence and Routing.}
Object-centric and geometry-aware representations have long been useful for robot learning and visuomotor control \citep{shridhar2022perceiver,zhu2022viola,chi2023diffusion,belkhale2023hydra}. Recent multimodal work extends this object-centric view to broader visual understanding, segmentation, editing, and generation settings \citep{ravi2024sam2,yuan2026lmms,zhong2026unified}. Recent visual-planning work further suggests that image-based intermediate reasoning can be better suited than text-space planning for spatial decisions \citep{xu2025visual}. Video and egocentric benchmarks similarly emphasize that object grounding should account for temporal context, reference granularity, and future state changes \citep{openx2023,yuan2025videorefer,yuan2025eoc}. As summarized in \Cref{tab:visual_cue_compare}, VISUALTHINK-VLA follows this direction by first constructing a six-channel candidate evidence bank and then deploying a compact four-channel operational interface for a frozen VLA backbone. The evidence channels are built from standard perception and vision-language back-ends such as Grounding DINO, SAM2, Qwen2.5-VL, CLIP, ViT, and OWL-ViT \citep{liu2023groundingdino,ravi2024sam2,qwen2025qwen25vl,radford2021learning,dosovitskiy2021image,minderer2022owlvit}. Region-level and spatio-temporal referring work further motivates preserving location and temporal cues as explicit evidence rather than compressing all visual information into a single global representation \citep{minderer2022owlvit,yuan2025pixelrefer}. Its routing mechanism is related to conditional computation, mixture-of-experts models, and dynamic vision-language expert tuning \citep{shazeer2017moe,lepikhin2020gshard,fedus2022switch,zhang2024hyperllava}, while the supervision and audit layer builds on faithfulness and rationale-evaluation work that emphasizes counterfactual validation over free-form explanations \citep{deyoung2020eraser,jacovi2020towards,siegel-etal-2024-probabilities,kamahi-yaghoobzadeh-2024-counterfactuals}.

\section{Method}
\label{sec:method}

\subsection{Method Overview}

\begin{figure}[!t]
    \centering
    \includegraphics[width=1\textwidth]{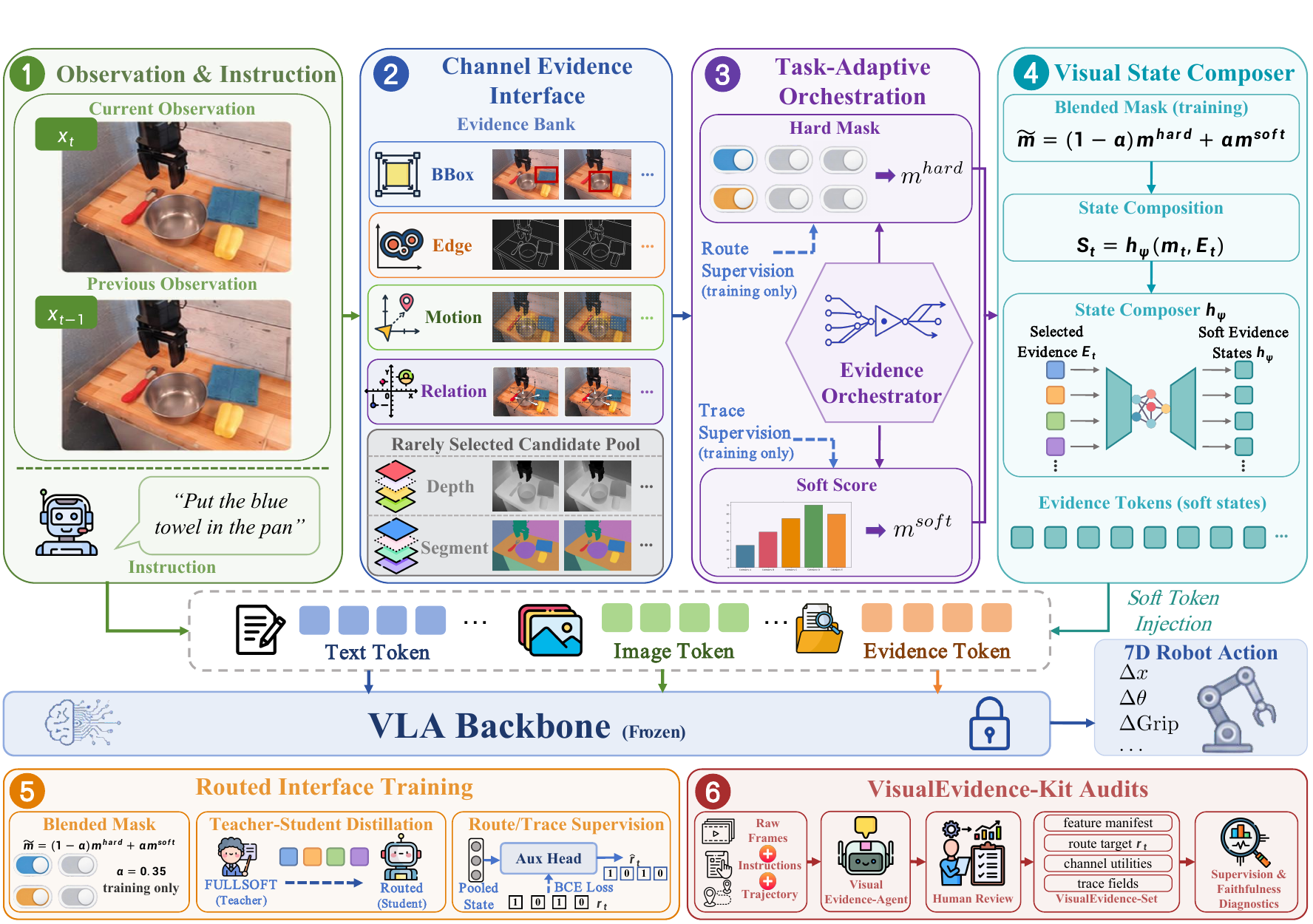}
    \caption{Overview of the \method{} pipeline. Dashed arrows denote training-only supervision.}
    \label{fig:method}
\end{figure}

The core design choice in \method{} is to avoid prompt-text evidence injection. Instead of expressing intermediate reasoning as free-form text, the frozen VLA backbone is conditioned on a small set of learned visual evidence states. As shown in \Cref{fig:method}, \method{} first constructs a six-channel candidate evidence bank, screens out two low-utility channels, and then uses a task-adaptive router over the retained four effective channels. The routed evidence is mapped into visual states through the Visual State Composer before action decoding. Under this formulation, \thinkimg{} means that action prediction is guided by routed image-grounded states rather than by a long textual trace.

\subsection{Channel Evidence Interface}
\label{subsec:soft_interface}

At decision step $t$, the policy receives the current RGB observation $x_t$, the previous observation $x_{t-1}$, and the language instruction $q$. From this context, we derive a six-channel candidate evidence bank
\begin{equation}
\mathcal{E}_t = \{e_t^{c} \mid c \in \mathcal{C}_{\mathrm{cand}}\},
\end{equation}
where
\begin{equation}
\begin{aligned}
\mathcal{C}_{\mathrm{cand}} = \{&
\texttt{bbox},\texttt{edge},\texttt{motion},\texttt{relation},\texttt{depth},\texttt{segment}\}.
\end{aligned}
\end{equation}
Each $e_t^c$ is a compact evidence vector. The construction depends on the observation pair and, for instruction-grounded channels, on $q$:
\begin{equation}
e_t^c = g_c(x_{t-1}, x_t, q).
\end{equation}
The six candidate channels cover complementary visual information: \texttt{bbox} encodes object location, \texttt{edge} captures boundary geometry, \texttt{motion} summarizes short-horizon change, \texttt{relation} represents instruction-grounded spatial relations, \texttt{depth} provides monocular geometry, and \texttt{segment} marks object regions. These channels are represented as compact vectors rather than image-sized tensors, so the VLA receives a lightweight visual interface instead of a dense auxiliary perception stack. This decomposition is aligned with recent object-centric and spatio-temporal grounding studies that separate object identity, region reference, and temporal change \citep{ravi2024sam2,yuan2025pixelrefer,yuan2025videorefer}. The detailed construction of each channel is described in the appendix.
However, channel screening identifies \texttt{depth} and \texttt{segment} as low-utility cues in our benchmark setting: they are rarely selected, bring marginal success gains, and increase side-perception overhead. We therefore define the operational evidence set and bank as
\begin{equation}
\begin{aligned}
\mathcal{C}_{\mathrm{op}} = \{&
\texttt{bbox},\texttt{edge},\texttt{motion},\texttt{relation}\}, \quad
\mathcal{E}_t^{\mathrm{op}}=\{e_t^c \mid c \in \mathcal{C}_{\mathrm{op}}\}.
\end{aligned}
\end{equation}
Accordingly, \textbf{\fullsoft{}} sets the route mask to $\mathbf{1}$ over $\mathcal{E}_t^{\mathrm{op}}$, while \textbf{\method{}} uses a router-selected sparse mask over the same four-channel bank.

\subsection{Task-Adaptive Orchestration}

A task-adaptive router predicts channel probabilities
\begin{equation}
m_t^{\mathrm{soft}} = r_{\phi}(x_{t-1}, x_t, q, \mathcal{E}_t^{\mathrm{op}}),
\end{equation}
and a hardening operator $H(\cdot)$ converts them into an inference-time route mask
\begin{equation}
m_t^{\mathrm{hard}} = H(m_t^{\mathrm{soft}}), \quad
m_t^{\mathrm{hard}} \in \{0,1\}^{|\mathcal{C}_{\mathrm{op}}|}.
\end{equation}
Given a mask $m$, we denote the routed evidence as
\begin{equation}
\mathcal{E}_t^{\mathrm{op}}(m)=\{m_{t,c}e_t^c \mid c \in \mathcal{C}_{\mathrm{op}}\}.
\end{equation}

At inference time, \method{} uses a single task-adaptive router across datasets. The hard mask $m_t^{\mathrm{hard}}$ activates only the evidence channels selected for the current decision. This is the main efficiency mechanism: all four operational channels are available to the interface, but the action decoder is exposed only to routed evidence states.

This separation also clarifies the relation among route variables. The predicted route is $m_t^{\mathrm{soft}}$, the inference-time selection is $m_t^{\mathrm{hard}}$, and the VisualEvidence-Set target $r_t$ used below is a supervised route label. The model is trained to align its predicted route with $r_t$, but at inference time it uses only its own router prediction.

\subsection{Visual State Composer}

The Visual State Composer is the concrete module denoted by $h_{\psi}$. It projects the routed channel vectors into a small set of learned visual states,
\begin{equation}
S_t = h_{\psi}(\mathcal{E}_t^{\mathrm{op}}(m)),
\end{equation}
and inserts them as conditioning states before action decoding. The frozen VLA backbone then predicts the action as
\begin{equation}
a_t = f_{\theta}(x_t, q, S_t), \quad \theta \ \text{frozen}.
\end{equation}
Because the composer consumes already-routed evidence, it adds only a lightweight learned interface; it does not invoke an online image-editing model or generate textual rationales at inference time.

\subsection{Training the Routed Interface}

Pure hard routing was brittle in difficult manipulation settings. During training, we therefore use a blended route mask
\begin{equation}
\bar m_t = (1-\alpha)m_t^{\mathrm{hard}} + \alpha m_t^{\mathrm{soft}},
\end{equation}
with $\alpha = 0.35$, and compute $S_t=h_{\psi}(\mathcal{E}_t^{\mathrm{op}}(\bar m_t))$. At inference time, $\bar m_t$ is replaced by $m_t^{\mathrm{hard}}$.

To transfer the capacity of dense evidence integration into the sparse route, we train the routed adapter with logits distillation from the four-channel \fullsoft{} teacher \citep{hinton2015distilling}. Let $p_{\mathrm{T}}^\tau$ and $p_{\mathrm{S}}^\tau$ denote teacher and student action-token distributions at temperature $\tau$. The dynamic loss is
\begin{equation}
\mathcal{L}_{\mathrm{dyn}}
= \mathcal{L}_{\mathrm{action}}
+ \lambda_{\mathrm{distill}}\tau^2
\mathrm{KL}(p_{\mathrm{T}}^\tau \Vert p_{\mathrm{S}}^\tau),
\end{equation}
with $\lambda_{\mathrm{distill}}=0.2$ and $\tau=1.5$.

VisualEvidence-Kit provides a supervised route target $r_t \in \{0,1\}^{|\mathcal{C}_{\mathrm{op}}|}$. A training-only auxiliary head predicts $\hat r_t$ from the pooled evidence state and route features. The final training objective is
\begin{equation}
\mathcal{L}_{\mathrm{total}}
= \mathcal{L}_{\mathrm{dyn}}
+ \lambda_{\mathrm{trace}}
\mathcal{L}_{\mathrm{BCE}}(\hat r_t, r_t).
\end{equation}
This route/trace supervision regularizes the evidence pathway during training; the auxiliary head and channel-grounded traces are not used at inference time.

\section{VisualEvidence-Kit}
\label{sec:visualevidence_kit}

To train and examine a policy that reasons through sparse visual evidence, we construct \textbf{VisualEvidence-Kit}, a route-grounded supervision and audit resource for VLA control. VisualEvidence-Kit contains two coupled components: a \textbf{VisualEvidence-Agent} that derives and governs evidence instructions from manipulation trajectories, and the resulting \textbf{VisualEvidence-Set} that supports route supervision and counterfactual auditing. Unlike ordinary action trajectories, its instructions explicitly describe what visual channels are relevant to the action decision and whether the routed policy depends on them. This design is also informed by data-centric learning work that uses adaptive pseudo-labeling and active annotation to handle imperfect supervision \citep{deyoung2020eraser,zhang2022boostmis}.

\begin{figure}[H]
    \centering
    \includegraphics[width=\columnwidth]{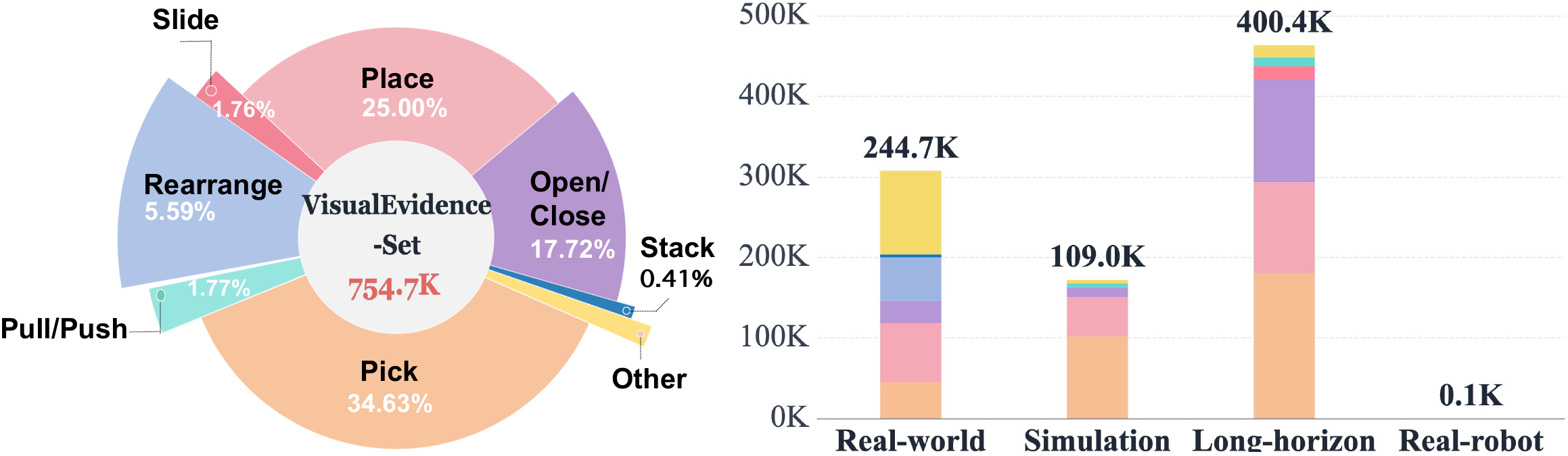}
    \caption{Data statistics of VisualEvidence-Set}
    \label{fig:dataset_information}
\end{figure}

\subsection{VisualEvidence-Agent}

The VisualEvidence-Agent converts raw frames and trajectory metadata into governed route-grounded instructions through four stages. 

\begin{figure}[!t]
    \centering
    \includegraphics[width=\textwidth]{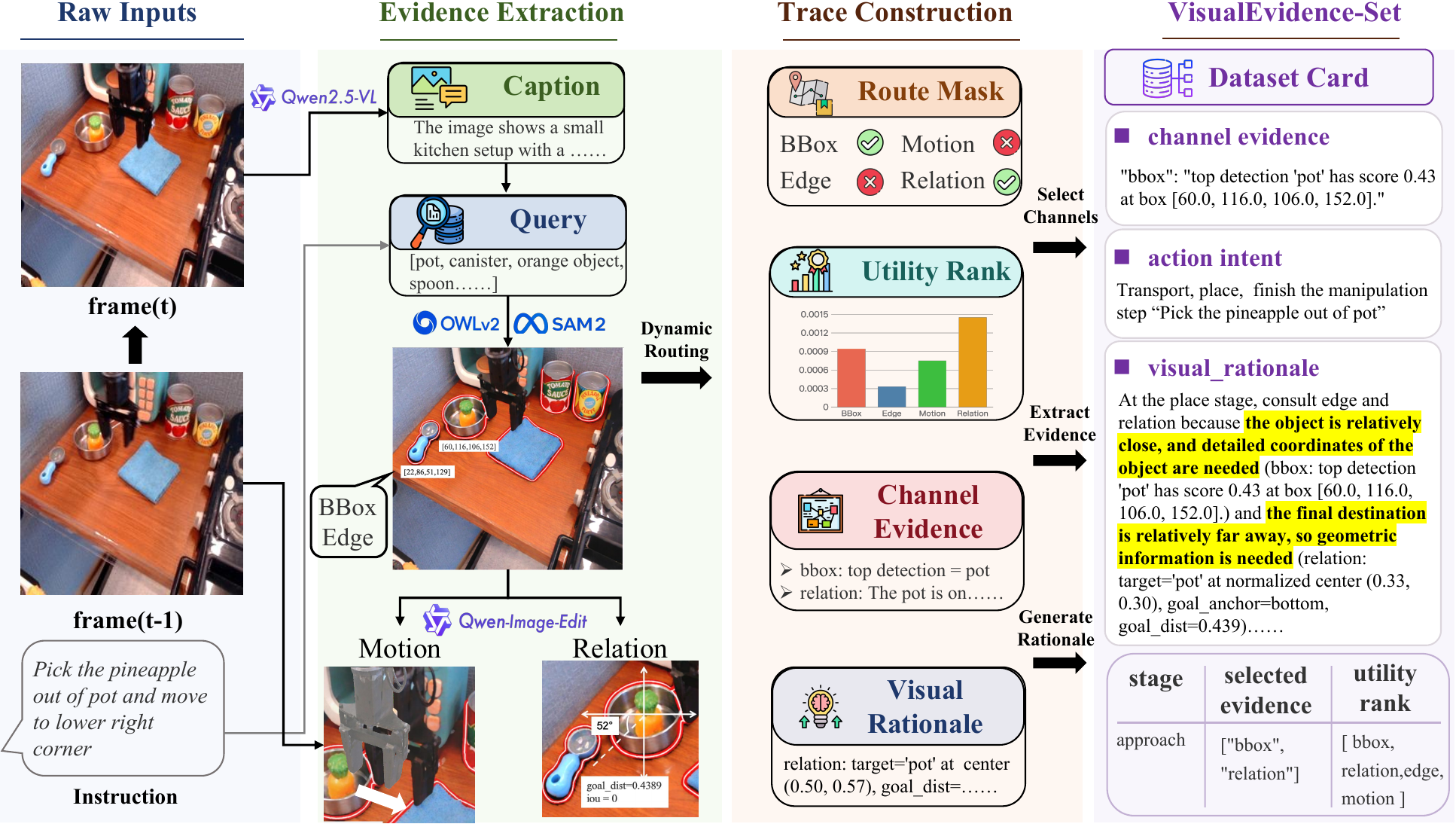}
    \caption{VisualEvidence-Kit workflow and VisualEvidence-Set record schema.}
    \label{fig:visualevidence_agent}
\end{figure}

\noindent\textit{(i) Evidence extraction}: applies the candidate channel extractors to construct a feature manifest for each decision context. 

\noindent\textit{(ii): Route and utility assessment}: aggregates routing signals and counterfactual channel utilities to form channel-level route targets and utility ranks. 

\noindent\textit{(iii): Trace construction}: records the manipulation stage, primitive, evidence dependence, difficulty, and selected evidence as a structured channel-grounded trace rather than a free-form rationale. 

\noindent\textit{(iv): Human review} verifies instruction consistency and filters unreliable labels before export. As illustrated in \Cref{fig:visualevidence_agent}, this filtering step follows the same practical concern as imperfect-label and uncertainty-aware learning: noisy supervision should be separated from high-confidence training or audit targets \citep{zhang2023learning,zhang2024revisiting}. This pipeline makes the supervision aligned with the visual interface used by \method{} while reserving the traces for training and audit only. 

\subsection{VisualEvidence-Set}

The resulting VisualEvidence-Set organizes heterogeneous manipulation information around the evidence demands of control. Each instruction contains the observation, instruction context, feature manifest, supervised route target $r_t$, counterfactual channel utilities, and channel-grounded trace fields. We further categorize the instructions along complementary axes, including scene domain, action category, manipulation stage, and routed visual evidence type, following the broader trend of heterogeneous knowledge organization in large vision-language systems \citep{qwen2025qwen25vl,lin2025healthgpt}. As summarized in \Cref{fig:dataset_information}, here displays the distribution of instructions for different action types and scene domains. In addition, each instruction is annotated by visual features, task stage, and difficulty level, enabling both router training and diagnostic evaluation across evidence requirements.

To support different use cases, we curate the constructed instructions into increasingly stringent subsets. \textit{Full-Clean} is used for corpus-level statistics and weighted training after basic consistency filtering; \textit{HQ-Trace} retains reliable structured traces for route- and trace-supervised refinement; and \textit{Gold-Faithfulness} serves as a high-reliability subset of 754.7k instructions for counterfactual audit experiments.

During policy training, VisualEvidence-Set provides route and trace supervision, while at inference time neither the agent nor its trace fields are required. For audit, we introduce counterfactual faithfulness diagnostics inspired by rationale-based and perturbation-based tests. These diagnostics evaluate whether \thinkimg{} is grounded in the policy's actually routed evidence rather than merely serving as a post-hoc explanation. The complete workflow and dataset statistics are detailed in Appendix~\ref{subsec:VisualEvidence-Kit_Workflow}.

\section{Experimental Setup}
\label{sec:experiments}

\noindent\textbf{Benchmark Roles.}
The multi-dataset control benchmark evaluates \method{} on \texttt{BridgeData V2}, \texttt{Fractal}, \texttt{RoboTurk}, \texttt{LIBERO}, and \texttt{UT Austin MUTEX} \citep{walke2023bridgedatav2,openx2023,mandlekar2018roboturk,liu2023libero,shah2023mutex}. VisualEvidence-Kit provides the route-grounded audit layer, and the real-robot study evaluates closed-loop deployment on a separate tabletop platform.

\noindent\textbf{Baselines.}
The external comparisons cover textual traces (ECoT), image-grounded reasoning (TraceVLA, SpatialVLA), and stronger reasoning-policy variants. Within our method family, we report a frozen \basevla{} baseline, a dense \fullsoft{} reference, and the sparse \method{} variant.

\noindent\textbf{Compared Systems.}
\textbf{\basevla{}-only} runs the frozen pretrained \openvla{} backbone, a 7B-parameter VLA model, with the original image and instruction inputs and no external evidence. \textbf{\fullsoft{}} enables all channels in the four-channel operational bank for every sample. \textbf{\method{}} uses a task-adaptive router to activate only a sparse subset of channels.

\noindent\textbf{Real-Robot Protocol.}
The real-robot study uses a desk-mounted \textbf{PIPER NERO (7F)} 7-DoF arm with a fixed external RGB camera \citep{kim24openvla,zawalski2024ecot,black2024pi0,black2025pi05,shukor2025smolvla}. The policy predicts end-effector-style actions, and a robot-side controller converts them into executable commands. The task suite covers \textbf{language-specified multi-object pick-place}, \textbf{relation-sensitive placement}, \textbf{contact-sensitive object reorientation}, and \textbf{two-stage compositional manipulation}.

\noindent\textbf{Metrics.}
The primary control metric is \texttt{success\_rate}. For efficiency, the main benchmark reports \texttt{avg\_step\_latency\_s}, measured as batch-1 wall-clock time from observation and instruction to action output after warm-up. We also report \texttt{avg\_completion\_time\_s} for real-robot evaluation and a route-grounded audit score for VisualEvidence-Kit.
\section{Results and Analysis}
\label{sec:results}


\begin{table}[H]
\centering
\scriptsize
\setlength{\tabcolsep}{2.8pt}
\renewcommand{\arraystretch}{1.22}
\resizebox{\textwidth}{!}{%
\begin{tabular}{p{0.18\textwidth}*{16}{c}}
\toprule
\textbf{Method} &
\multicolumn{6}{c}{\hdr{\textbf{Real-World Scenes}}} &
\multicolumn{6}{c}{\hdr{\textbf{Simulation Scenes}}} &
\multicolumn{4}{c}{\hdr{\textbf{Long-Horizon Tasks}}} \\
\cmidrule(lr){2-7}\cmidrule(lr){8-13}\cmidrule(lr){14-17}
&
\multicolumn{2}{c}{\textbf{BridgeData V2}} &
\multicolumn{2}{c}{\textbf{Fractal}} &
\multicolumn{2}{c}{\textbf{RoboTurk}} &
\multicolumn{2}{c}{\textbf{LIBERO-Object}} &
\multicolumn{2}{c}{\textbf{LIBERO-Goal}} &
\multicolumn{2}{c}{\textbf{LIBERO-Spatial}} &
\multicolumn{2}{c}{\textbf{LIBERO-Long}} &
\multicolumn{2}{c}{\textbf{UT Austin MUTEX}} \\
\cmidrule(lr){2-3}\cmidrule(lr){4-5}\cmidrule(lr){6-7}\cmidrule(lr){8-9}\cmidrule(lr){10-11}\cmidrule(lr){12-13}\cmidrule(lr){14-15}\cmidrule(lr){16-17}
& \textbf{Succ. (\%) $\uparrow$} & \textbf{Lat. (s) $\downarrow$} & \textbf{Succ. (\%) $\uparrow$} & \textbf{Lat. (s) $\downarrow$} & \textbf{Succ. (\%) $\uparrow$} & \textbf{Lat. (s) $\downarrow$} & \textbf{Succ. (\%) $\uparrow$} & \textbf{Lat. (s) $\downarrow$} & \textbf{Succ. (\%) $\uparrow$} & \textbf{Lat. (s) $\downarrow$} & \textbf{Succ. (\%) $\uparrow$} & \textbf{Lat. (s) $\downarrow$} & \textbf{Succ. (\%) $\uparrow$} & \textbf{Lat. (s) $\downarrow$} & \textbf{Succ. (\%) $\uparrow$} & \textbf{Lat. (s) $\downarrow$} \\
\midrule
\rowcolor{tblgrouptext}
\multicolumn{17}{c}{\textit{Textual CoT / autoregressive reasoning before action}} \\
\textbf{ECoT}
& 85.09 & 8.377 & 32.87 & 6.702 & 15.78 & 6.397
& - & - & - & - & - & -
& - & - & 20.32 & 6.255 \\
\midrule
\rowcolor{tblgrouprl}
\multicolumn{17}{c}{\textit{Reinforced or hybrid thinking-action decoding}} \\
\textbf{DeepThinkVLA-RL}
& 27.93 & 1.204 & 33.67 & 1.219 & 2.12 & 1.323
& \textbf{98.58} & 1.359 & 96.32 & 1.277 & 93.97 & 1.224
& 95.02 & 1.202 & 26.09 & 0.772 \\
\midrule
\rowcolor{tblgroupvisual}
\multicolumn{17}{c}{\textit{Spatial or image-grounded thinking}} \\
\textbf{TraceVLA}
& 86.87 & 0.404 & 86.25 & 0.402 & 93.27 & 0.416
& 85.68 & 0.425 & 75.24 & 0.407 & 85.90 & 0.428
& 52.62 & 0.440 & 1.10 & \textbf{0.403} \\
\textbf{SpatialVLA}
& 86.57 & 0.589 & 67.85 & 0.582 & 91.50 & 0.585
& 87.50 & 0.481 & 79.38 & 0.479 & 86.86 & 0.490
& 53.25 & 0.498 & 0.65 & 0.544 \\
\textbf{InternVLA-M1}
& 11.18 & 1.153 & 23.46 & 1.153 & 8.51 & 1.151
& 96.04 & 1.021 & 92.52 & 1.021 & 92.26 & 1.022
& 90.65 & 1.021 & 12.37 & 1.144 \\
\midrule
\rowcolor{tblgroupmatched}
\multicolumn{17}{c}{\textit{OpenVLA-family re-evaluation}} \\
\textbf{\basevla{}} 
& 75.37 & 0.345 & 87.45 & 0.345 & 95.56 & 0.342
& 89.30 & 0.343 & 80.83 & 0.343 & 84.60 & 0.342
& 54.86 & 0.302 & 41.09 & 0.349 \\
\textbf{\fullsoft{}}
& 88.45 & 0.447 & 90.38 & 0.448 & 96.32 & 0.445
& 97.78 & 0.445 & \textbf{97.14} & 0.445 & 96.39 & 0.445
& \textbf{96.24} & 0.497 & 77.10 & 0.551 \\
\rowcolor{tblours}
\textbf{\method{}}
& \textbf{89.49} & \textbf{0.367} & \textbf{90.82} & \textbf{0.367} & \textbf{96.10} & \textbf{0.415}
& 97.74 & \textbf{0.385} & 97.05 & \textbf{0.345} & \textbf{96.69} & \textbf{0.356}
& 95.87 & \textbf{0.421} & \textbf{77.26} & 0.451 \\
\bottomrule
\end{tabular}
}
\caption{Performance comparison under different datasets. $\uparrow$ indicate higher-is-better metrics, and $\downarrow$ indicate lower-is-better metrics.}
\label{tab:benchmark_suite}
\end{table}

\noindent
\textbf{Primary Control Comparison.}
\Cref{tab:benchmark_suite} shows that \method{} achieves a favorable accuracy-latency trade-off across diverse control settings. Compared with the matched \basevla{} re-evaluation, \method{} improves success on seven of eight benchmarks, with the largest gains on BridgeData V2, the LIBERO family, and UT Austin MUTEX. It also reduces BridgeData V2 latency from 8.377\,s with ECoT to 0.367\,s while improving success. Compared with the dense \fullsoft{} teacher, \method{} preserves comparable accuracy but reduces latency on every benchmark, suggesting that sparse routed evidence retains most dense-evidence benefits while avoiding unnecessary channel exposure.

\begin{figure}[!t]
    \centering
    \includegraphics[width=\textwidth]{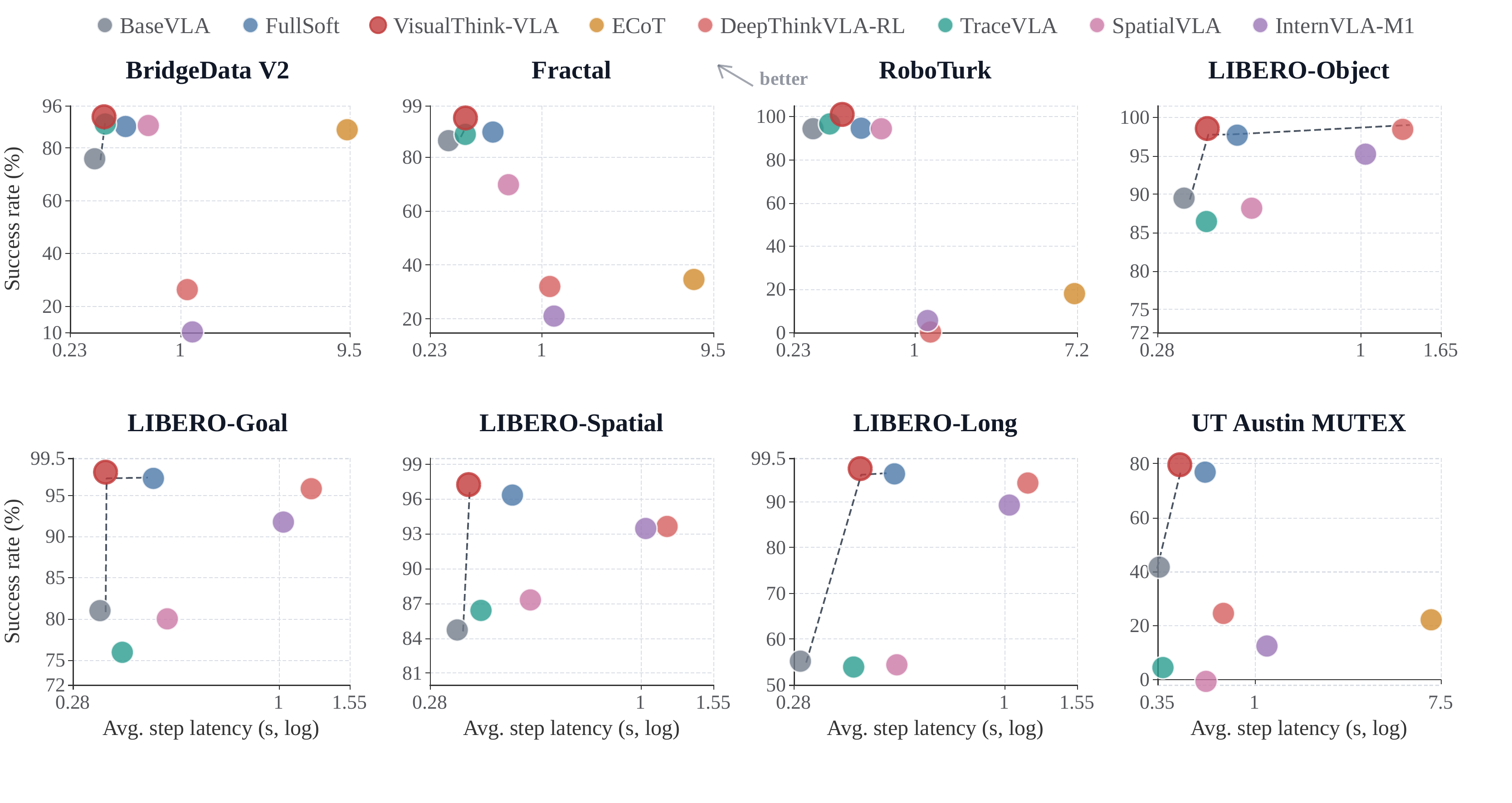}
    \caption{Success-latency trade-off across the eight main benchmarks.}
    \label{fig:benchmark_tradeoff}
\end{figure}

\Cref{fig:benchmark_tradeoff} visualizes the same suite as success-latency frontiers. It makes the empirical role of \method{} clearer than success alone: textual reasoning baselines can improve selected tasks but remain far outside the sub-second control regime, while the sparse routed interface stays close to the low-latency OpenVLA-family region and preserves most of the gains of dense visual evidence.

\noindent\textbf{Backbone Portability.}
\method{} is designed as a routed visual reasoning layer rather than an \openvla{}-specific modification. To test this point, \Cref{tab:backbone_portability} evaluates the same evidence interface on the VisualEvidence-Set test split with three representative VLA base policies that differ in pretraining recipe, action parameterization, and baseline latency.

\begin{table}[H]
\centering
\scriptsize
\setlength{\tabcolsep}{4pt}
\renewcommand{\arraystretch}{1.18}
\resizebox{0.8\textwidth}{!}{%
\begin{tabular}{p{0.22\textwidth}cccccc}
\toprule
\textbf{Backbone} &
\multicolumn{2}{c}{\hdr{\textbf{Base policy}}} &
\multicolumn{2}{c}{\hdr{\textbf{+\method{}}}} &
\multicolumn{2}{c}{\hdr{\textbf{Change}}} \\
\cmidrule(lr){2-3}\cmidrule(lr){4-5}\cmidrule(lr){6-7}
& \textbf{Succ. (\%)} & \textbf{Lat. (s)} & \textbf{Succ. (\%)} & \textbf{Lat. (s)} & \textbf{$\Delta$Succ.} & \textbf{$\Delta$Lat.} \\
\midrule
\openvla{} \citep{kim24openvla} & 76.26 & 0.339 & 92.63 & 0.388 & +16.37 & +0.050 \\
Octo \citep{octo2024} &  49.52 & 0.606 & 60.39 & 0.683 & +10.87 & +0.077 \\
SmolVLA \citep{shukor2025smolvla} & 42.73 & 0.248 & 54.68 & 0.352 & +11.95 & +0.104 \\
\bottomrule
\end{tabular}
}
\caption{Backbone portability study on the VisualEvidence-Set test split across three representative VLA base policies.}
\label{tab:backbone_portability}
\end{table}

\Cref{tab:backbone_portability} shows that the test-set gains are preserved across all three backbones with only modest latency increases. This supports the interpretation of \method{} as a general visual reasoning layer for VLA policies rather than an architecture-specific modification that only works with \openvla{}.


\begin{table}[H]
\centering
\scriptsize
\setlength{\tabcolsep}{3.0pt}
\renewcommand{\arraystretch}{1.18}
\resizebox{\textwidth}{!}{%
\begin{tabular}{p{0.17\textwidth}p{0.16\textwidth}*{12}{c}}
\toprule
\textbf{Variant} & \textbf{Evidence path} &
\multicolumn{6}{c}{\hdr{\textbf{Real-World Scenes}}} &
\multicolumn{4}{c}{\hdr{\textbf{Long-Horizon \& Simulation Tasks}}} &
\multicolumn{2}{c}{\hdr{\textbf{Average}}} \\
\cmidrule(lr){3-8}\cmidrule(lr){9-12}\cmidrule(lr){13-14}
&
& \multicolumn{2}{c}{\textbf{BridgeData V2}}
& \multicolumn{2}{c}{\textbf{Fractal}}
& \multicolumn{2}{c}{\textbf{RoboTurk}}
& \multicolumn{2}{c}{\textbf{LIBERO family}}
& \multicolumn{2}{c}{\textbf{UT Austin MUTEX}}
& \textbf{Succ. (\%) $\uparrow$} & \textbf{Lat. (s) $\downarrow$} \\
\cmidrule(lr){3-4}\cmidrule(lr){5-6}\cmidrule(lr){7-8}\cmidrule(lr){9-10}\cmidrule(lr){11-12}
& & \textbf{Succ. (\%) $\uparrow$} & \textbf{Lat. (s) $\downarrow$} & \textbf{Succ. (\%) $\uparrow$} & \textbf{Lat. (s) $\downarrow$} & \textbf{Succ. (\%) $\uparrow$} & \textbf{Lat. (s) $\downarrow$} & \textbf{Succ. (\%) $\uparrow$} & \textbf{Lat. (s) $\downarrow$} & \textbf{Succ. (\%) $\uparrow$} & \textbf{Lat. (s) $\downarrow$} & & \\
\midrule
\textbf{\basevla{}-only} & none
& 75.37 & 0.345 & 87.45 & 0.345 & \textbf{96.56} & 0.342 & 77.40 & 0.333 & 41.09 & 0.349 & 75.57 & 0.343 \\
Prompt-text evidence & prompt text
& 87.49 & 1.384 & 86.82 & 1.389 & 94.90 & 1.448 & 95.07 & 1.453 & 71.76 & 1.467 & 87.21 & 1.428 \\
Heavy dense & dense side cue
& 86.69 & 0.536 & 88.22 & 0.542 & 93.20 & 0.607 & 91.67 & 0.613 & 72.71 & 0.663 & 86.50 & 0.592 \\
\textbf{\fullsoft{}} & dense soft tokens
& 88.45 & 0.447 & 90.38 & 0.448 & 96.32 & 0.445 & \textbf{96.89} & 0.458 & 77.10 & 0.551 & 89.83 & 0.470 \\
\rowcolor{tblours}
\textbf{\method{}} & routed soft tokens
& \textbf{89.49} & 0.367 & \textbf{90.82} & 0.367 & \textbf{96.10} & 0.415 & 96.84 & 0.377 & \textbf{77.26} & 0.451 & \textbf{90.10} & 0.395 \\
\bottomrule
\end{tabular}
}
\caption{Internal interface comparison. $\uparrow$ indicate higher-is-better metrics, and $\downarrow$ indicate lower-is-better metrics.}
\label{tab:baseline_comparison}
\end{table}

\noindent\textbf{Internal Interface Comparison.}
\Cref{tab:baseline_comparison} shows that the gains come from the structured evidence interface rather than simply adding auxiliary signals. Prompt-text evidence improves over \basevla{} but remains slow due to autoregressive token decoding. Heavy dense evidence uses all six candidate channels and incurs higher latency with lower average success. In contrast, \method{} improves over \fullsoft{} in average success (90.10\% vs.\ 89.83\%) while reducing latency from 0.470\,s to 0.395\,s.

\begin{figure}[H]
    \centering
    \includegraphics[width=\textwidth]{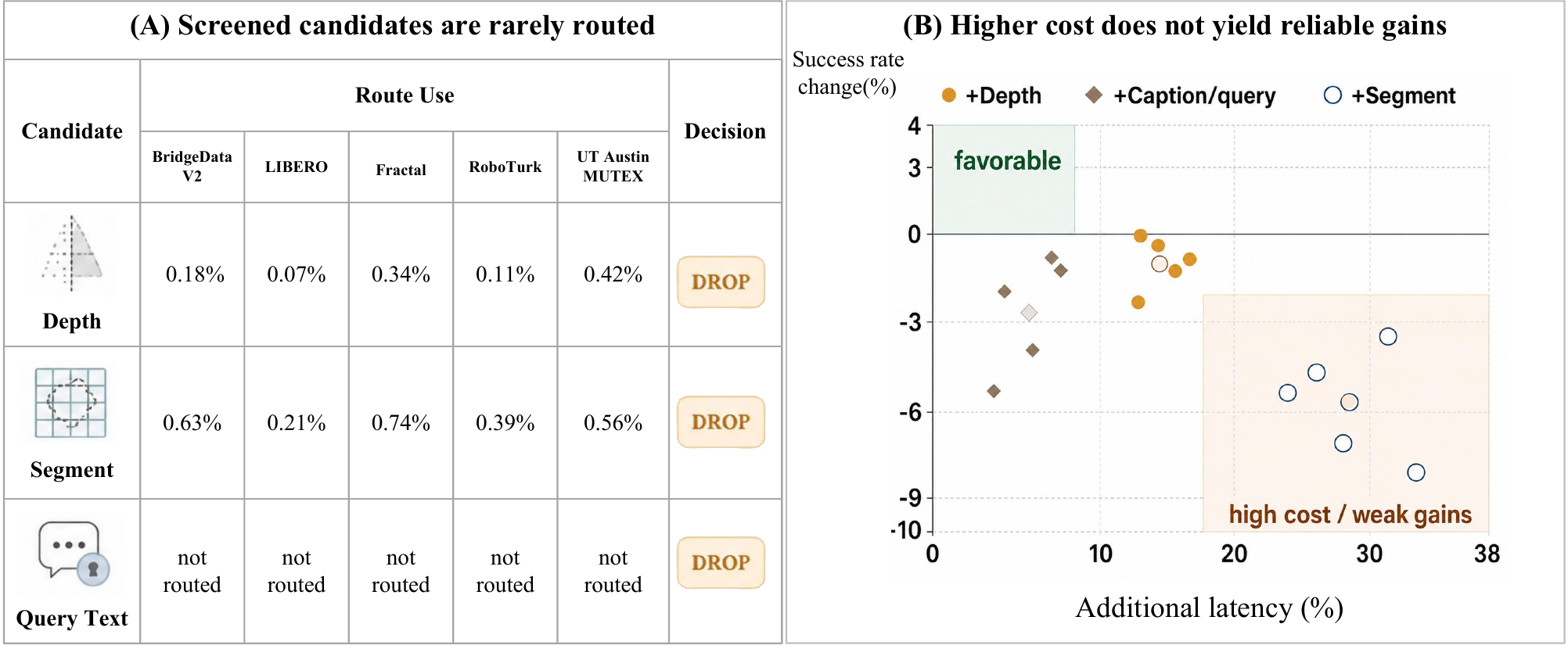}
    \caption{Evidence-channel screening for the retained routed evidence bank.}
    \label{fig:channel_screening}
\end{figure}

The screening results in \Cref{fig:channel_screening} further show that the initial six-channel bank can be reduced to a four-channel operational subset without losing the main control gains. Dense screened candidates do not justify their extra runtime cost, and prompt-text evidence partially recovers performance but departs from visual intermediate reasoning.


\begin{table}[H]
\centering
\scriptsize
\setlength{\tabcolsep}{3pt}
\resizebox{\textwidth}{!}{%
\begin{tabular}{p{0.24\textwidth}p{0.16\textwidth}p{0.16\textwidth}ccc}
\toprule
\textbf{Task family} & \textbf{Evidence} & \textbf{Metric} & \textbf{\basevla{}} & \textbf{\fullsoft{}} & \textbf{\method{}} \\
\midrule
Multi-object pick-place & \texttt{bbox} + distractors & Success (\%) $\uparrow$ & 48.6 & 73.4 & \textbf{75.6} \\
Relation-sensitive placement & \texttt{relation} & Success (\%) $\uparrow$ & 45.6 & 63.7 & \textbf{67.2} \\
Contact-sensitive reorientation & \texttt{edge}/\texttt{motion} & Success (\%) $\uparrow$ & 63.4 & \textbf{85.8} & 83.5 \\
Two-stage compositional task & stage-wise routing & Success (\%) $\uparrow$ & 26.5 & 57.4 & \textbf{59.2} \\
All tasks & efficiency & Avg. completion time (s) $\downarrow$ & 22.7 & 30.2 & 25.6 \\
All tasks & route sparsity & Avg. selected ch. & 0 & 4.0 & 1.83 \\
\bottomrule
\end{tabular}
}
\caption{Focused real-robot closed-loop evaluation on the four-task tabletop suite. Each task family is evaluated over 50 trials. Up arrows indicate higher-is-better metrics, and down arrows indicate lower-is-better metrics.}
\label{tab:real_robot}
\end{table}

\noindent\textbf{Real-Robot Evaluation.}
\Cref{tab:real_robot} shows that the trade-off transfers to live closed-loop control. \method{} outperforms \basevla{} on all four task families and improves over \fullsoft{} on three of them. The only exception is contact-sensitive reorientation, where always-on four-channel dense evidence is slightly stronger. Nevertheless, \method{} remains more efficient than \fullsoft{} in completion time (25.6\,s vs.\ 30.2\,s) while selecting only 1.83 channels on average.


\noindent
\textbf{VisualEvidence-Kit Analysis.}
VisualEvidence-Kit tests whether routed evidence is inspectable and behaviorally relevant. On the held-out audit set, \method{} achieves the strongest route-grounded audit score (0.840) and utility overlap (0.984), while prompt-text evidence reaches 0.391 and exposes no visual route. Stage-wise routing is also sensitive to task phase: \texttt{bbox} dominates during \texttt{approach}, \texttt{motion} rises sharply at \texttt{grasp}, and \texttt{edge} remains consistently active.

\begin{figure}[H]
    \centering
    \includegraphics[width=\textwidth]{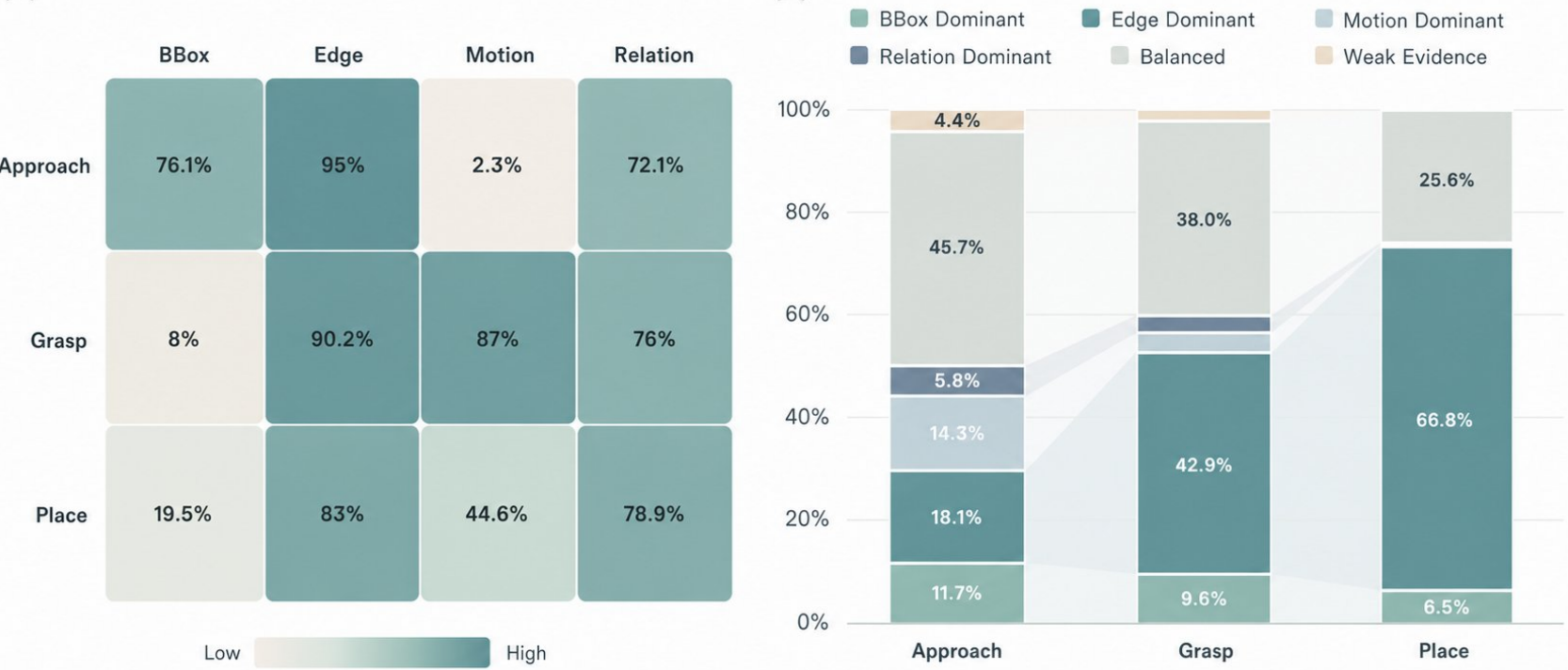}
    \caption{Before-version post-hoc stage-wise routing analysis on VisualEvidence-Set.}
    \label{fig:routing_sparsity}
\end{figure}

\Cref{fig:routing_sparsity} reports the full stage-conditioned routing pattern behind these audit results. The figure supports the same conclusion as the aggregate scores: the router does not merely select a fixed shortcut channel, but changes evidence use with the phase of manipulation.

\section{Ablation Studies}
\label{sec:ablations}


\begin{figure}[H]
\centering
\includegraphics[width=\textwidth]{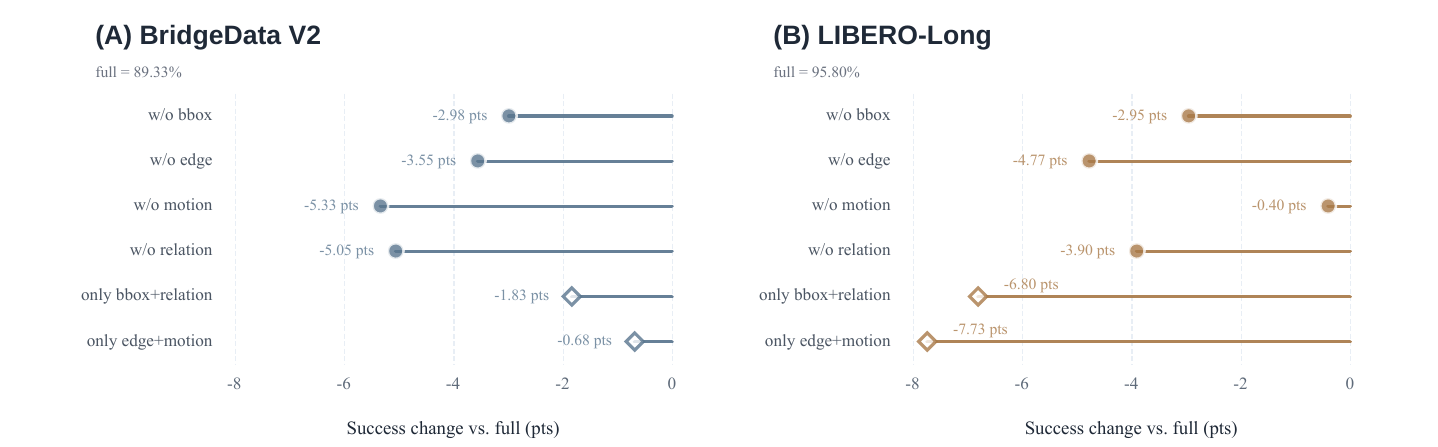}
\caption{Feature ablations on BridgeData V2 and LIBERO-Long. The before-version plot reports success change relative to the full retained-channel model; w/o denotes removing the listed evidence channel or channel pair.}
\label{fig:feature_ablation}
\end{figure}

\noindent\textbf{Feature Ablations.}
\Cref{fig:feature_ablation} evaluates the retained four-channel model on BridgeData V2 and LIBERO-Long, which stress cluttered manipulation and long-horizon control. In this before-version ablation plot, removing each retained channel reduces success relative to the full model, and the magnitude of the drop is task-dependent. BridgeData V2 is more sensitive to \texttt{bbox}, \texttt{motion}, and \texttt{relation}, whereas LIBERO-Long shows the largest drops when removing \texttt{edge} or using only narrow channel pairs. This indicates that the routed evidence bank is complementary rather than dominated by a single cue.

\noindent\textbf{Orchestration Ablations.}
Orchestration ablations show that the final sparse routing recipe improves both trainability and efficiency.

\begin{figure}[H]
\centering
\includegraphics[width=0.86\textwidth]{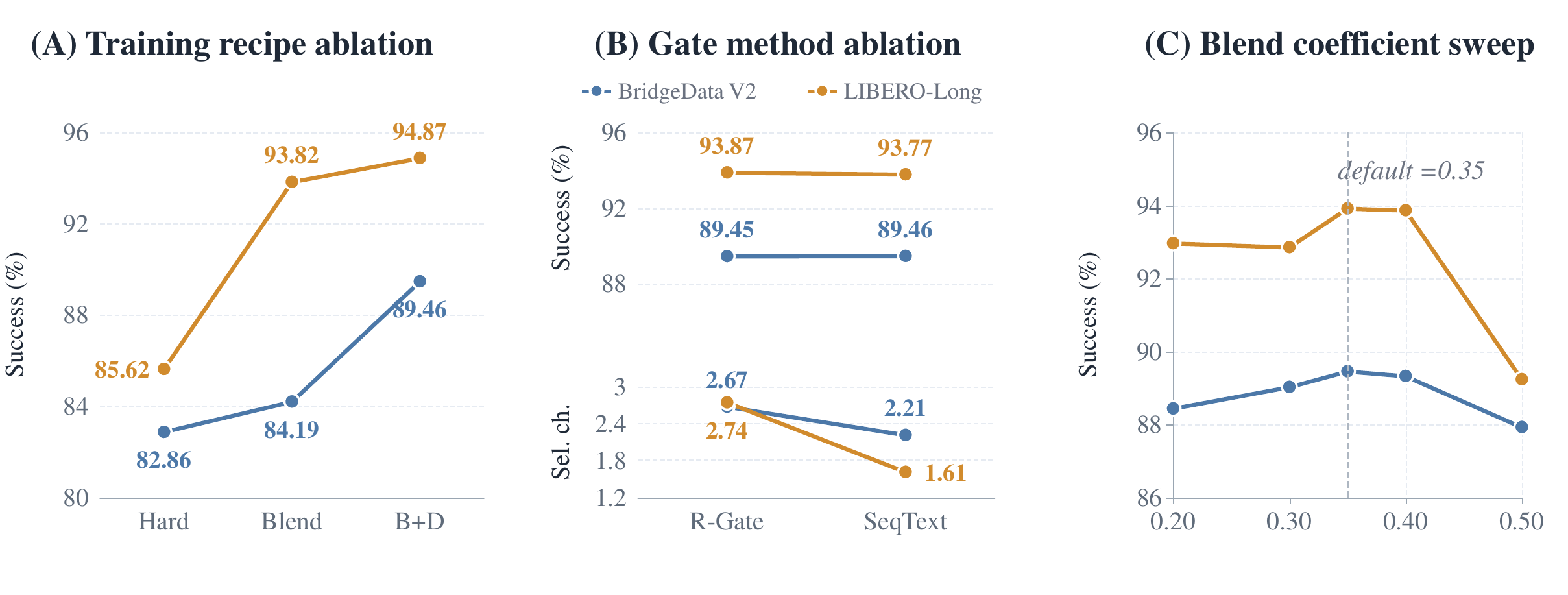}
\caption{Evidence-orchestration ablations on BridgeData V2 and LIBERO-Long, where B+D denotes blend plus distillation, R-Gate denotes routed evidence gate, SeqText denotes the sequence-text gate baseline, and Sel. ch. denotes the average number of selected evidence channels.}
\label{fig:recipe_ablation}
\end{figure}

\Cref{fig:recipe_ablation} separates the effects of training recipe, gate design, and blend coefficient. Soft-hard collaborative routing improves over hard sparse masks, increasing success from 82.86\% to 84.19\% on BridgeData V2 and from 85.62\% to 93.82\% on LIBERO-Long. Adding teacher-student optimization further raises success to 89.46\% and 94.87\%, respectively. The task-adaptive router then maintains comparable success to the sequence-text gate while selecting fewer channels on average, supporting the efficiency of the final orchestration design.

\noindent\textbf{Route-Trace Ablations.}
\begin{multicols}{2}
Route and trace supervision mainly improve faithfulness rather than raw success. Removing route supervision reduces route alignment from 0.929 to 0.758, while removing trace supervision reduces utility mention from 0.984 to 0.738, with only moderate success drops in both cases. Replacing the channel-grounded target with a free-form rationale baseline causes the strongest degradation, lowering route alignment to 0.052 and utility mention to 0.039.

\columnbreak

\noindent\begin{minipage}{\columnwidth}
\centering
\scriptsize
\setlength{\tabcolsep}{2pt}
\resizebox{\linewidth}{!}{%
\begin{tabular}{@{}lcccc@{}}
\toprule
\textbf{Variant} & \textbf{Success} & \textbf{Route aln.} & \textbf{Util. ment.} & \textbf{$\Delta$TopRemove} \\
\midrule
\method{} & \textbf{90.46}\% & \textbf{0.929} & \textbf{0.984} & - \\
w/o route sup. & 87.40\% & 0.758 & 0.742 & -3.06 pts \\
w/o trace sup. & 87.33\% & 0.882 & 0.738 & -3.13 pts \\
w/o utility rank & 88.47\% & 0.847 & 0.793 & -1.99 pts \\
free-form tgt. & 82.43\% & 0.052 & 0.039 & -4.97 pts \\
\bottomrule
\end{tabular}
}
\captionof{table}{Interpretability ablations on the VisualEvidence-Set trace-control slice. Here, \emph{Route aln.} denotes route alignment, \emph{Util. ment.} denotes utility mention, and $\Delta$TopRemove denotes the performance change after removing the top-ranked routed channel.}
\label{tab:trace_ablation_appendix}
\end{minipage}
\end{multicols}

Together, \Cref{tab:trace_ablation_appendix} indicates that inspectable intermediate reasoning in VLA control benefits from structured, channel-grounded supervision rather than unconstrained explanatory text.

\section{Conclusion}

\method{} introduces a sparse image-grounded evidence interface for frozen VLA policies. In the \openvla{} instantiation studied here, the routed variant preserves much of the benefit of dense soft evidence while reducing control-time overhead, and it remains effective in a real-robot closed-loop setting.
The results indicate that, for embodied control, compact structured evidence in visual space is a more practical intermediate reasoning substrate than long textual reasoning traces.

\section{Limitations}
\label{sec:limitations}

 The current implementation focuses on visual channels that are broadly useful for manipulation, while other signals such as tactile feedback, force sensing, audio, or longer-term memory are left for future extensions. The evaluation spans several public benchmarks and a focused real-robot tabletop suite; broader deployments across more embodiments, workspaces, and longer-horizon tasks would further test the generality of the routed evidence interface. Finally, as with other robot-learning policies, real-world deployment should be paired with appropriate monitoring and task-specific safety constraints.

\beginappendix

\section*{Appendix}
\label{sec:appendix}

This appendix provides supporting implementation and evaluation details. We document the VisualEvidence-Set governance pipeline, describe the evidence-channel construction, spell out the real-robot task suite, and collect qualitative and audit details.

\section{VisualEvidence-Kit Workflow}
\label{subsec:VisualEvidence-Kit_Workflow}

The VisualEvidence-Agent builds the VisualEvidence-Set in a staged manner: it first derives the candidate evidence channels from raw observations, then aggregates route targets and utility signals, and finally sends the generated records to a manual review and annotation stage before exporting the governed benchmark entries with channel-grounded trace fields.

\begin{figure}[!t]
    \centering
    \includegraphics[width=\columnwidth]{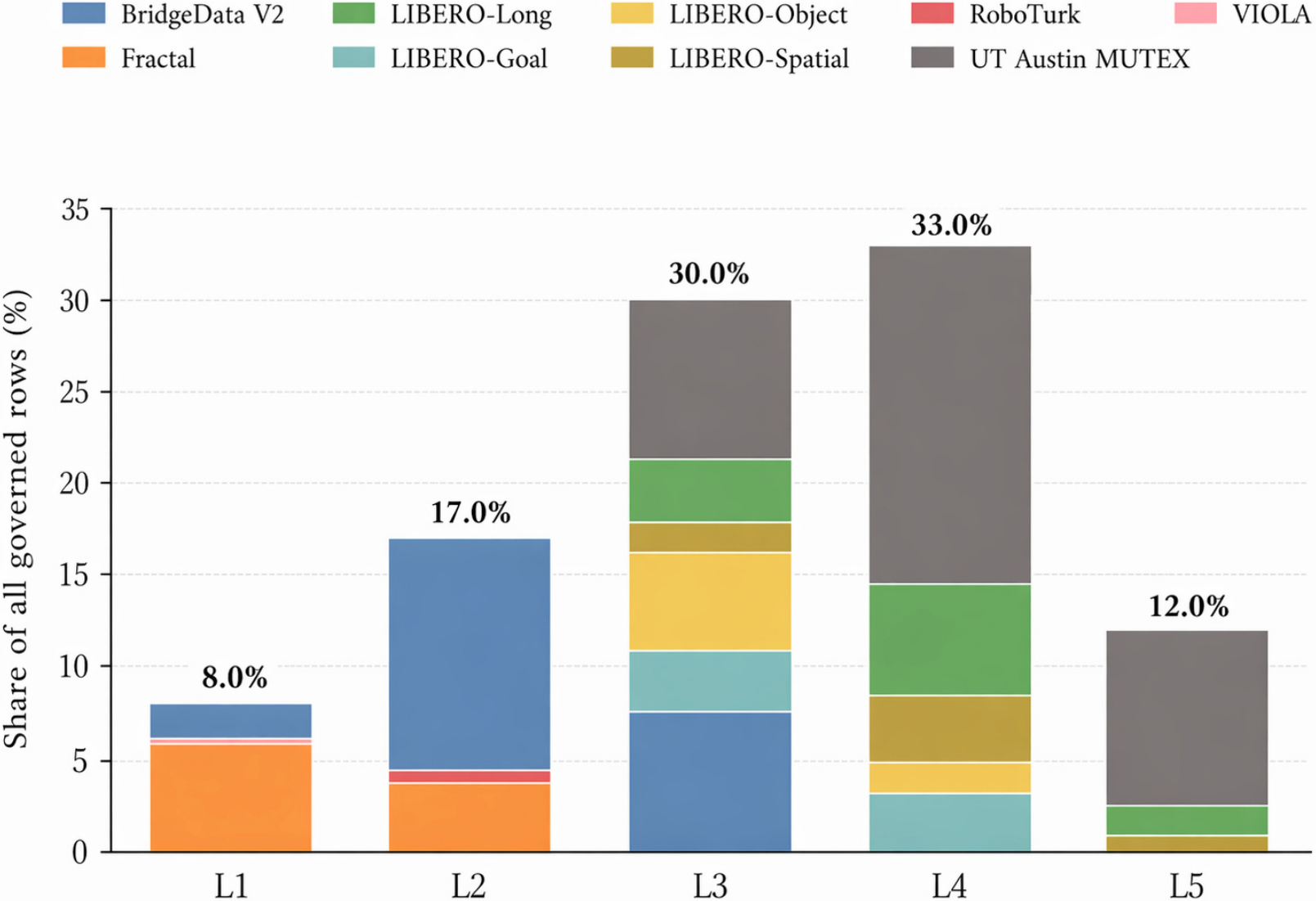}
    \caption{VisualEvidence-Set governance pipeline for sorting raw traces into progressively stricter subsets.}
    \label{fig:data_sort}
\end{figure}

\begin{table}[!t]
\centering
\scriptsize
\setlength{\tabcolsep}{4pt}
\renewcommand{\arraystretch}{1.18}
\resizebox{0.8\textwidth}{!}{%
\begin{tabular}{lrrrrr}
\toprule
\textbf{Source} & \textbf{Raw traces} & \textbf{Full-Clean} & \textbf{HQ-Trace} & \textbf{Gold-Faithfulness} & \textbf{Review/drop} \\
\midrule
BridgeData V2 & 210,622 & 203,948 & 188,736 & 146,680 & 15,212 \\
Fractal & 89,675 & 86,104 & 78,442 & 52,318 & 7,662 \\
LIBERO family & 273,465 & 271,982 & 267,541 & 253,806 & 4,441 \\
RoboTurk & 7,127 & 6,842 & 6,091 & 4,376 & 751 \\
UT Austin MUTEX & 361,883 & 354,726 & 336,918 & 297,462 & 17,808 \\
VIOLA & 64 & 61 & 55 & 48 & 6 \\
\bottomrule
\end{tabular}
}
\caption{VisualEvidence-Set governance readout. Full-Clean is used for broad statistics and weighted training, HQ-Trace for trace-supervised refinement, and Gold-Faithfulness for high-confidence audit experiments.}
\label{tab:data_governance}
\end{table}

\Cref{fig:data_sort,tab:data_governance} summarize how raw route-grounded traces are governed before use. Full-Clean removes invalid or inconsistent records and supports broad dataset statistics and weighted training. HQ-Trace keeps higher-quality channel-grounded traces for trace-supervised refinement. Gold-Faithfulness is the strictest subset and is used for high-confidence counterfactual audit experiments.

\paragraph{Evidence-channel construction.}
The six-channel candidate bank is extracted offline before policy inference. \textbf{\texttt{bbox}} uses open-vocabulary detection with Grounding DINO / OWL-ViT-style localization to encode coarse object location and target extent \citep{liu2023groundingdino,minderer2022owlvit}. \textbf{\texttt{edge}} uses SAM2 mask proposals and boundary rasterization to capture object contours and local shape discontinuities \citep{ravi2024sam2}. \textbf{\texttt{motion}} uses short-horizon frame differencing over $(x_{t-1}, x_t)$ together with SAM2-based temporal mask propagation and lightweight code-based motion rendering to capture immediate scene change \citep{ravi2024sam2}. \textbf{\texttt{relation}} encodes instruction-grounded spatial geometry with Qwen2.5-VL parsing and deterministic relation rendering, such as left-of, in-front-of, or inside relations \citep{qwen2025qwen25vl}. \textbf{\texttt{depth}} uses Depth Anything V2 to estimate dense monocular depth and pools it into a compact geometry cue \citep{yang2024depthanythingv2}. \textbf{\texttt{segment}} uses detector-prompted SAM2 masks to encode object regions and coarse scene partitioning \citep{liu2023groundingdino,ravi2024sam2}. At policy inference time, \method{} consumes only the compact evidence vectors and the router-selected soft states; no image-editing model is invoked online.

\section{Real-Robot Task Suite}
\label{subsec:Real-Robot Task Suite}

The main paper reports only the aggregate real-robot result table because the primary point is to compare deployment-time behavior across \basevla{}, \fullsoft{}, and \method{}. This section spells out the tabletop task families behind that summary. The task design follows the mechanism of the method: one family stresses target localization under distractors, one isolates instruction-grounded spatial relations, one emphasizes contact-sensitive pose change, and one tests stage-dependent evidence usage within a short compositional task. \Cref{tab:real_robot_tasks} lists representative instructions for each family.

The deployment platform is a desk-mounted \textbf{PIPER NERO (7F)} 7-DoF arm with \textbf{1.5\,kg} payload, \textbf{580\,mm} reach, and \textbf{$\pm 0.1$\,mm} repeatability. The arm is controlled from a host PC through the vendor \textbf{CAN} interface, while RGB observations are captured by a fixed external camera.
    
\begin{table}[H]
\centering
\scriptsize
\setlength{\tabcolsep}{10pt}
\begin{tabular}{@{}ll@{}}
\toprule
\textbf{Task family} & \textbf{Representative instructions} \\
\midrule
Multi-object pick-place & \makecell[l]{Pick the block from the cluttered area and place it into the bowl.} \\
Relation-sensitive placement & \makecell[l]{Move the block to the left side of the cup and stop there.} \\
Contact-sensitive reorientation & \makecell[l]{Flip the mug upright and release it once the base is stable.} \\
Two-stage compositional manipulation & \makecell[l]{Move the block to the drawer area, then open the drawer and finish placement.} \\
\bottomrule
\end{tabular}
\caption{Real-robot task definitions for the tabletop deployment study. Each task family lists two representative instructions; the detailed stressors and success criteria are described in the surrounding text.}
\label{tab:real_robot_tasks}
\end{table}

These task definitions make the real-robot section readable without relying on a large qualitative montage.

\section{Qualitative Case Studies}
\label{subsec:Qualitative Case Studies}

\begin{figure}[!t]
    \centering
    \includegraphics[width=\textwidth]{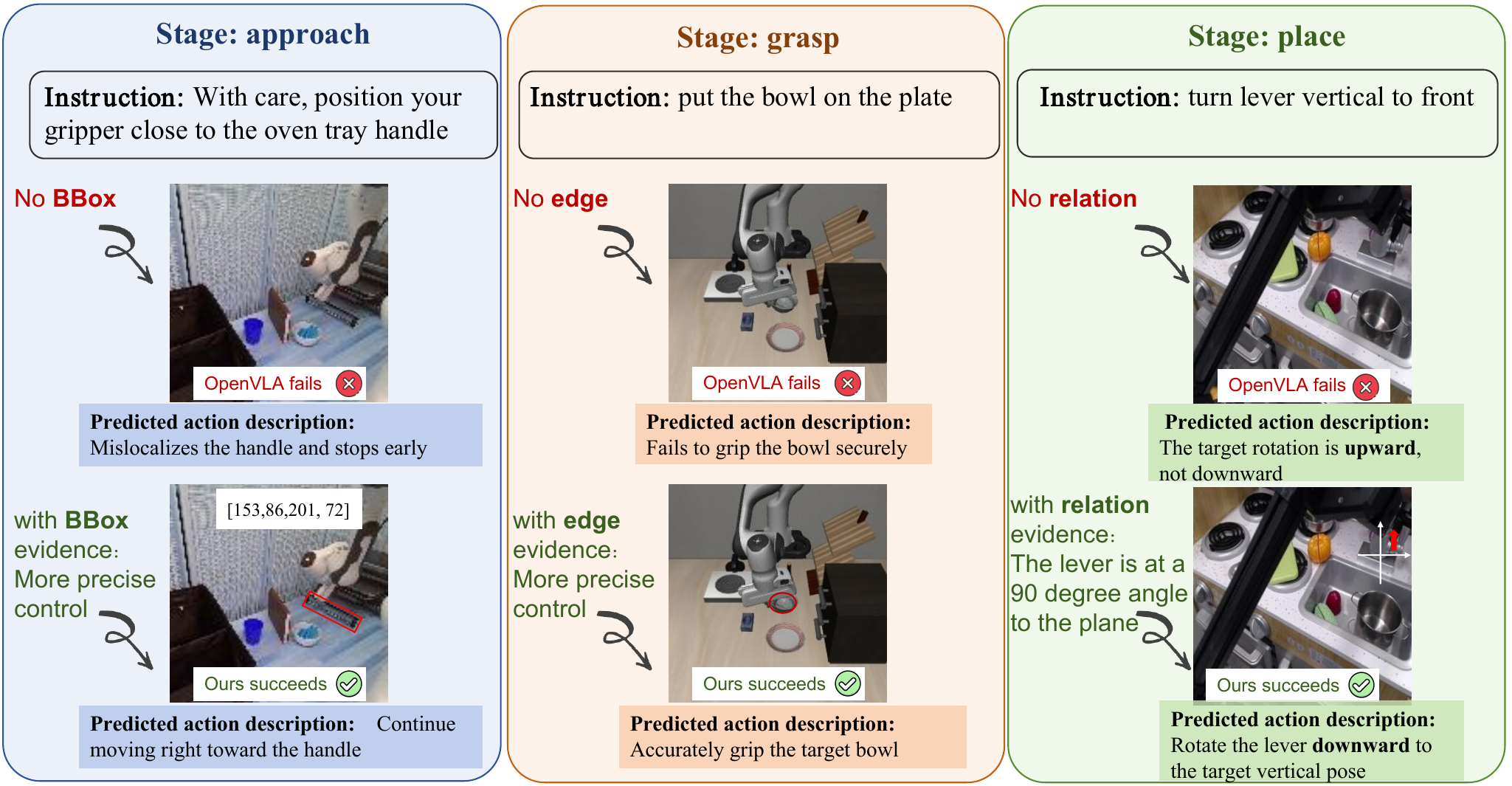}
    \caption{Qualitative case studies of routed visual evidence in representative manipulation tasks.}
    \label{fig:case_study}
\end{figure}

\Cref{fig:case_study} provides qualitative examples of the routed visual evidence used in representative closed-loop cases.

The case studies illustrate how routed visual evidence changes action-relevant perception rather than merely adding decoration. In the lever-rotation example, the relation cue makes the target pose explicit: without relation evidence, the base policy rotates in the wrong direction, while the routed policy identifies the lever's orientation relative to the plane and selects the correct downward rotation. In the bowl-placement example, edge evidence sharpens the contact and grasp boundary, helping the policy close on the bowl securely instead of producing an unstable grasp. In the oven-tray example, bbox evidence localizes the handle under visual clutter, preventing early stopping caused by target mislocalization. Across the three cases, the selected channels correspond to the active task demand: relation for pose and spatial orientation, edge for precise contact geometry, and bbox for target localization.

\section{VisualEvidence-Kit Audit and Faithfulness Details}
\label{subsec:VisualEvidence-Kit Audit and Faithfulness Details}

\begin{multicols}{2}
The main text positions VisualEvidence-Kit as a supervision-and-audit layer rather than another control benchmark. VisualEvidence-Agent builds the VisualEvidence-Set used for the trace-quality summary in \Cref{tab:evidence_trace}, complementing the stage-conditioned routing visualization in the main text.

\columnbreak

\noindent\begin{minipage}{\columnwidth}
\centering
\scriptsize
\setlength{\tabcolsep}{3pt}
\resizebox{\linewidth}{!}{%
\begin{tabular}{@{}lcccc@{}}
\toprule
\textbf{Method} & \textbf{Trace} & \textbf{Route} & \textbf{Util.} & \textbf{Sel. ch.} \\
\midrule
Prompt-text evidence & 0.391 & \routeno{} & 0.768 & 4.00 \\
Heavy dense perception & 0.441 & \routeyes{} & 0.755 & 2.24 \\
\fullsoft{} & 0.791 & \routeyes{} & 0.961 & 4.00 \\
\method{} & \textbf{0.840} & \routeyes{} & \textbf{0.984} & 2.22 \\
\bottomrule
\end{tabular}
}
\captionof{table}{VisualEvidence-Kit supervision and audit summary. Here, \emph{Trace} denotes the channel-grounded trace alignment score, \emph{Route} indicates whether explicit route information is exposed, \emph{Util.} denotes utility overlap, and \emph{Sel. ch.} denotes the average number of selected channels.}
\label{tab:evidence_trace}
\end{minipage}
\end{multicols}

Taken together with the routing figure in the main text, \Cref{tab:evidence_trace} gives a fuller account of the audit layer than any single number alone. It summarizes whether the produced traces expose explicit route information and whether the stated evidence overlaps with the most useful channels. The governance taxonomy uses four compact axes: stage, primitive, evidence dependence, and difficulty.

\section{Human Review Protocol}

Human review in VisualEvidence-Kit is used only to check route-grounded records, evidence-channel labels, and trace quality. Reviewers were laboratory students familiar with the robot manipulation setting. They were given task-level guidelines describing the meaning of each evidence channel, the expected route labels, and the intended research use of the reviewed records for VLA supervision, audit, and diagnostic evaluation. No crowdworking platform was used and no monetary payment was provided.

The review process did not collect personal data from reviewers or from external human subjects. The reviewed content consists of robot observations, task instructions, generated evidence labels, and audit fields. Reviewers were informed that the reviewed records would be used for research dataset construction and evaluation. Because the work does not involve human-subject behavioral data, personal information, or intervention with human participants, no separate ethics-board protocol was sought.

\section{Artifact Licenses}

We use external datasets, VLA backbones, and perception tools only under their original release conditions. The public robot-learning benchmarks used for evaluation, including BridgeData V2, Open X-Embodiment/Fractal, RoboTurk, LIBERO, and UT Austin MUTEX, are cited in the main experimental section and remain governed by their respective dataset licenses and terms of use. The VLA backbones and comparison systems, including \openvla{}, Octo, SmolVLA, and related reasoning-augmented VLA methods, are used as research artifacts according to the conditions specified by their original authors. The perception back-ends used to construct visual evidence channels, including Grounding DINO, OWL-ViT, SAM2, Qwen2.5-VL, and Depth Anything V2, are likewise cited and used under their original model or code licenses.

Our use of these artifacts is consistent with their intended research use. External datasets are used for benchmark evaluation, VLA backbones are used for policy comparison and adaptation, and perception models are used to derive auxiliary visual evidence for research analysis. VisualEvidence-Kit and VisualEvidence-Set are intended only for VLA supervision, audit, and diagnostic evaluation. Any derivatives should remain within the access conditions of the original datasets and models from which they are derived.

The data used and created in this work consists of robot manipulation observations, task instructions, routed evidence labels, and audit records. We do not collect personal conversations, user profiles, human-subject records, or social-media text, and the VisualEvidence-Set does not contain information intended to name or uniquely identify individual people. We also do not introduce offensive-content annotations or generate offensive textual content as part of the dataset.

\section{AI Assistant Use}

We used AI assistants, including Codex and GPT-5.5, to support manuscript editing, language polishing, and local engineering assistance during paper preparation.

\bibliographystyle{unsrtnat}
\bibliography{main}

\end{document}